\documentclass[runningheads]{llncs}

\usepackage{graphicx}
\usepackage{hyperref}

\usepackage{amsmath,amssymb}
\usepackage{color}

\usepackage[width=122mm,left=12mm,paperwidth=146mm,height=193mm,top=12mm,paperheight=217mm]{geometry}

\usepackage{subfig}
\usepackage{amsmath}

\DeclareMathOperator*{\argmax}{arg\,max}

\usepackage{fancyhdr}
\usepackage{ifthen}

\newcommand{\version}[1]
{
 \ifthenelse{\equal{#1}{ARXIV}}{\thispagestyle{fancy}}{}
 \ifthenelse{\equal{#1}{WORKSHOP}}{\thispagestyle{plain}}{}
} 

\begin{document}

\def\ECCV18SubNumber{***}  

\title{PanoRoom: From the Sphere to the 3D Layout} 
\titlerunning{PanoRoom: From the Sphere to the 3D Layout}

\authorrunning{C. Fernandez, J.M. F\'acil, A. Perez, C. Demonceaux and J.J. Guerrero}

\author{Clara Fernandez-Labrador$^{1,2}$, Jos\'e M. F\'acil$^{1}$, Alejandro Perez-Yus$^{1}$, C\'edric Demonceaux$^{2}$, and Jose J. Guerrero$^{1}$}

\institute{$^{1}$Instituto de Investigaci\'on en Ingenier\'ia de Arag\'on (I3A), Universidad de Zaragoza, Spain. {\tt\small \{cfernandez, jmfacil, alperez, josechu.guerrero\} @unizar.es}. \\
$^{2}$ Le2i VIBOT ERL-CNRS 6000, Universit\'e de Bourgogne Franche-Comt\'e, France {\tt\small \{cedric.demonceaux@ubfc.fr\}}}

\maketitle

\version{ARXIV} % ARXIV OR WORKSHOP
\pagestyle{plain}

\begin{abstract}
We propose a novel FCN able to work with omnidirectional images that outputs accurate probability maps representing the main structure of indoor scenes, which is able to generalize on different data. Our approach handles occlusions and recovers complex shaped rooms more faithful to the actual shape of the real scenes. We outperform the state of the art not only in accuracy of the 3D models but also in speed.
\end{abstract}

\section{Introduction}

The problem of recovering the 3D layout of a cluttered indoor scene goes back to the early days of computer vision and still is a core research topic, as it is a key technology in several emerging application markets like augmented and virtual reality, indoor navigation, SLAM and robotics in general.
While witnessing the rapid progress on layout recovery methods from perspective images with both geometry and deep learning techniques, the expansion to omnidirectional vision is yet limited. 
Panoramic images have broken the barriers of performance on this task. 
PanoContext \cite{PanoContext} was the first to extend the frameworks designed for perspective images to panoramas. They recover both the layout, assumed as a 3D box (4 walls), and bounding boxes of the main objects inside the room. 
With the motivation to leave behind the assumption of simple boxes, \cite{fernandez2018layouts} generates layout hypotheses by geometric reasoning from structural corners obtained through geometry and deep learning combination. Most recent work is LayoutNet \cite{zou2018layoutnet}, which generates 3D layout models by training a network with panoramas and vanishing lines to obtain edge and corner maps. 

Here, we propose a Fully Convolutional Neural Network (FCN) that faces the 3D layout recovery problem from panoramas with the following contributions: first, we introduce a fully convolutional architecture using ResNet-50 \cite{he2016deep}, pretrained on ImageNet, as backbone. We include a single decoder that jointly predicts edge and corner maps, which requires fewer parameters and computing time than the state of the art while gaining accuracy. Notwithstanding that the great majority of the images are labeled as 4-walls rooms, our network predicts room solutions without assuming box-type rooms any more, from which we produce geometrically consistent room layouts. Our 3D models outperform existing methods on standard benchmark datasets \cite{Xiao2012,2017arXiv170201105A}.

\section{Deep learning structure perception}

\begin{figure}[t!]
\centering
\subfloat{\includegraphics[width=0.93\linewidth]{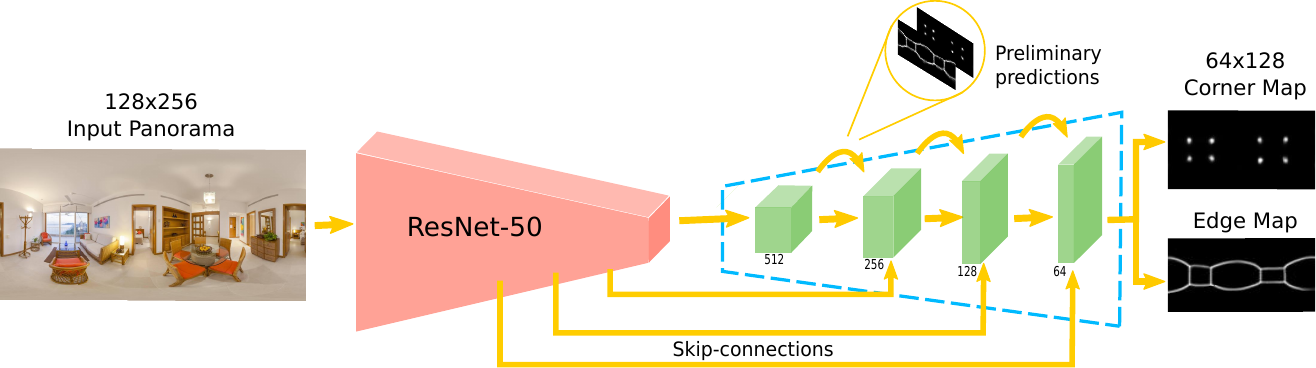}}
\caption{PanoRoom architecture: from ResNet-50, we build a single decoder that jointly predicts edge and corner maps already refined}
\label{fig:pipelineNet}
\end{figure}

The proposed FCN follows the encoder-decoder structure and builds upon ResNet-50 \cite{he2016deep}. We replace the final fully-connected layer with a decoder that jointly predicts layout edges and corners locations already refined. We illustrate the proposed architecture in Fig. \ref{fig:pipelineNet}. 
Our ground truth (GT) for every panorama consists of two probability maps. One represents the room edges whereas the other encodes the corner locations. Every pixel has a value of 1 if it contains structural information or 0 if it is background. Line thickening and Gaussian blur are both employed to make the model training easier to converge since the natural distribution of pixels is unbalanced, {\em e.g.} 95\% is background. 
Instead, similar approaches usually need to divide the GT into different classes, \cite{Mallya:2015,zhao2017physics,RoomNet}.

\textbf{Encoder.}
Most of deep-learning approaches facing layout recovery problem have made use of the VGG16 as encoder \cite{Mallya:2015,RoomNet}. Instead, \cite{zhao2017physics} builds their model over ResNet-101 \cite{he2016deep} outperforming the state of the art. Here, we propose ResNet-50 \cite{he2016deep}, pre-trained on the ImageNet dataset, which leads to a faster convergence due to the general low-level features learned from ImageNet. Residual networks allow us to increase depth without increasing the number of parameters with respect to their plain counterparts. This leads, in ResNet-50, to capture a receptive field of $483 \times 483$, enough for our input resolution  $128 \times 256$.  

\textbf{Decoder.}
Most recent work \cite{Mallya:2015,zou2018layoutnet} build two output branches for multi-task learning which leads to more computation time and more parameters. We instead propose a unique branch whose output has two channels, corners and edges maps, which helps to reinforce the quality of both map types.
In the decoder, we combine two different ideas. First, skip-connections \cite{ronneberger2015u} from the encoder to the decoder. Specifically, we concatenate `up-convolved' features with their corresponding features from the contracting part.  Second, we perform preliminary predictions in different resolutions which we also concatenate and feed back to the network following the spirit of \cite{dosovitskiy2015flownet}, see Fig. \ref{fig:pipelineNet}. 

\textbf{Loss functions.} 
Edge and corner maps are learned through a pixel-wise sigmoid cross-entropy loss function. Since we know a priori that these maps have an extremely unbalanced distribution of edge and corner labels, we introduce the ponder factors $\lambda_1$ and $\lambda_0$. Where 1 and 0 are the positive and negative classes respectively and $\lambda_c = \frac{N}{N_c}$, being $N$ the total number of pixels and $N_c$ amount of pixels of class $c$. The total sigmoid cross-entropy loss is the mean of the per-pixel loss: 
$\mathcal{L}_{Mi}  = 
 \lambda_1 \big( y_i\big(-\log(S(\hat{y}_i))\big)\big) + 
 \lambda_0 \big( (1-y_i) \big(-\log(1-S(\hat{y}_i))\big) \big)$ for every pixel $i$, where $y$ is the GT, $\hat{y}$ is the predicted map and $S$ is the sigmoid function.
Inspired by \cite{johnson2016perceptual}, we also define a perceptual loss function that measure high-level perceptual differences between images. We make use of an auto-encoder with same structure as the main network and we train this auto-encoder to encode the GT. Apart from encouraging the output images, $\hat{y}$, to exactly match the target images, $y$, we also encourage them to have similar feature representations. Hence, having $y$ as GT and $\mathcal{I}$ as input image $\mathcal{L}_{P}(y,\mathcal{I})  = ||\hat\phi_{j}(\mathcal{I}) - \phi_{j}(y)||^2_{2}$, where $\phi_{j}$ is the feature map on the $j^{th}$ hidden layer for the auto-encoder and $\hat\phi_{j}$ on our network.  
\begin{figure}[t!]
\centering
\subfloat{\includegraphics[width=0.95\linewidth]{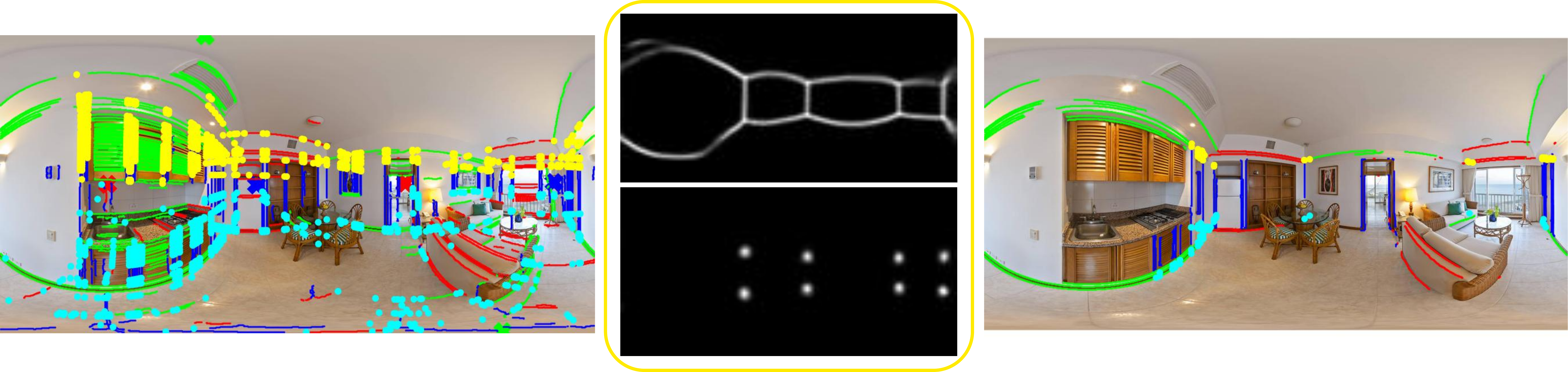}}
\caption{\textbf{Left:} Vanishing lines and corners obtained only with geometric reasoning. \textbf{Center:} Edge and Corner Maps predicted by our FCN. \textbf{Right:} Vanishing lines and corners after combining both approaches.}
\label{fig:linepr}
\end{figure}

\section{Layout recovery}
While deep-learning approaches have shown tremendous success and provide a deeper understanding of the scene, their output alone is insufficient as it does not enforce geometric constraints and priors. For this reason, we take advantage of our FCN output to produce geometrically consistent room layouts by optimizing over the deep learning clues under Manhattan World assumption, whereby there exist three orthogonal main directions that define the scene. Here, we do not assume the scene as a 4 wall box any more.
Lines and vanishing points (VP) extraction in perspective images has been satisfactorily faced so that many proposals working with panoramas \cite{PanoContext,yang2016efficient} sample perspective subviews to make use of them, with the subsequent increase of computation time. Recently, some approaches \cite{bazin2012globally,fernandez2018layouts} proposed methods to obtain lines and VP directly on panoramas improving the overall efficiency of the method. Here, we use the RANSAC-approach of \cite{fernandez2018layouts} that has demonstrate also to be faster than other methods.
Each extracted line is associated to a probability given by the sum of probabilities of the pixels it occupies in the edge map. This allows to remove those lines with 0 probability, leading to an optimal subset of accurate lines that allow our proposal to obtain final results with very few layout hypotheses generated. 
To make the most of our network output, we intersect the structural lines to obtain candidate corners that are scored with the corner map. See Fig. \ref{fig:linepr}.
The layout generation process follows the idea of \cite{fernandez2018layouts} but we use a different approximation to select the best hypothesis solution. Let's define each layout hypothesis as a combination of its edges and corners retrieved, $L^{h}( L^{h}_{E},L^{h}_{C})$. We associate them with the probability sum of the pixels they occupy in the corresponding predicted edge and corner maps, we define the per-pixel probability as $P_{edge}$ and $P_{corner}$ respectively. In this manner, we select the layout that maximizes the match between these two sources of information, $L^{Best} = \argmax (w_{e} \sum P_{edge}(L^{h}_{E})+ w_{c} \sum P_{corner}(L^{h}_{C}))$, where $w_{e}$ and $w_{c}$ are the term weights used to give them same importance.

\section{Experiments}

Experiments of the proposed approach are conducted within two public datasets that comprise several indoor scenes, SUN360 \cite{Xiao2012} and Stanford (2D-3D-S) \cite{2017arXiv170201105A}.
We take advantage of the $\thicksim$ 500 panoramas from the SUN360 dataset labeled by \cite{PanoContext} but, since panoramas were all labeled as box-type rooms, we hand label and substitute 35 panoramas representing more faithfully the actual shapes of the rooms. We split the raw dataset consisting in 85$\%$ training and 15$\%$ test scenes. 
For experiments with the Stanford 2D-3D dataset, we use same testing images (area 5) and GT provided by \cite{zou2018layoutnet}.

The input to the network is a single panoramic image of resolution 128$\times$256, unlike \cite{zou2018layoutnet} that uses also vanishing lines as input. The outputs, edge and corner maps, have resolution 64$\times$128. 
We apply horizontal mirroring as well as horizontal rotation from 0$^\circ$ to 360$^\circ$ of input images during training as data augmentation. 
We minimized the cross-entropy loss using Adam, regularized by penalizing the loss with the sum of the L2 of all weights. Initial learning rate is $2.5e^{-4}$ and exponentially decayed by a rate of 0.995 every epoch. We apply 0.3 dropout rate and 5e-6 weight decay.
Although we label some panoramas accurately to their actual shape, we still have a big unbalanced dataset. In order to overcome this problem, we choose a batch size of 16 and we force it always to include one example between those panoramas hand labeled by us (not box-type). This favors the learning of more complex rooms despite having few examples. 
The network is implemented in TensorFlow and trained with NVIDIA Geforce GTX 1080. 

\begin{figure}[t!]
\centering
\subfloat{\includegraphics[width=0.97\linewidth]{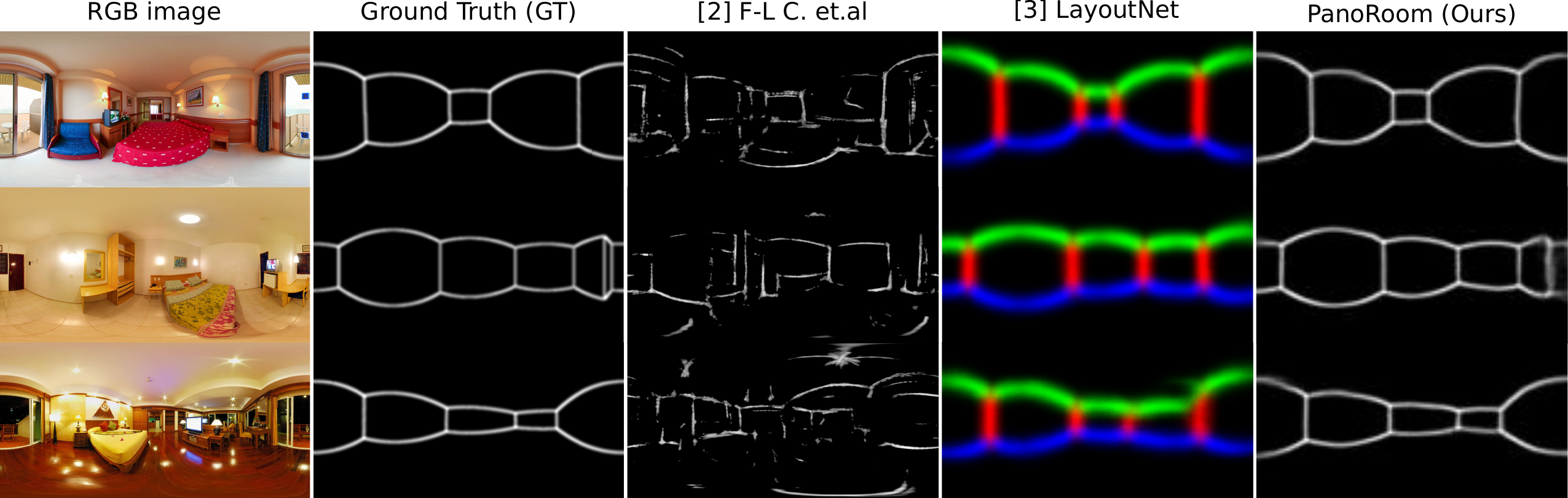}}
\caption{\textbf{Predicted Edge Maps}. Our FCN outperforms last works from the state of the art \cite{fernandez2018layouts,zou2018layoutnet}. We predict cleaner edges around the boundaries and recover faithful edge maps even when indoor scenes are not simple 4-walls rooms}
\label{fig:fcnnresults}
\end{figure}

\textbf{FCN Evaluation.}
After training our network on the SUN360 dataset, we first carried out an experiment to choose more precisely the weights leading to the best performance. To do that, we evaluate the predicted maps with their corresponding GT at different stages of the training on the same dataset. We saw that with 200 epochs (1 epoch = 25 iterations), our network reaches the best performance with a training time of $\thicksim$ 1 hour and a half. Testing takes about 0.7s per image. These results are collected in Table \ref{table:PRA2}, first row. Additionally, we find interesting to also evaluate results on the Stanford 2D-3D dataset to see how well our FCN is able to generalize, this results are shown in the second row. As it was expected, results are not as good when testing with a different dataset. However, we demonstrate in next experiments that it is enough to get layout reconstructions with little error and thus, our network is able to generalize.
\setlength{\tabcolsep}{4pt}
\begin{table}[t!]
\begin{center}
\caption{Evaluation of our FCN trained on SUN360 dataset for 200 epochs}
\label{table:PRA2}
\scriptsize 
\begin{tabular}{ccccc}
\hline\noalign{\smallskip}
\textbf{Dataset} &\textbf{ Precision(\%)} &\textbf{Recall(\%)}&\textbf{F1 Score(\%)}&\textbf{Accuracy(\%)}\\
\noalign{\smallskip}
\hline
\noalign{\smallskip}
SUN360 & 0.8338 &  0.8564& 0.8449& 0.9615\\ 
Stnfd.2D-3D & 0.6309 & 0.5904& 0.6113& 0.9160\\ 

\hline
\end{tabular}
\end{center}
\end{table}
\setlength{\tabcolsep}{1.4pt}
In Fig. \ref{fig:fcnnresults} we show 3 examples of our predicted maps with different number of walls compared to the GT and to other two approaches \cite{fernandez2018layouts,zou2018layoutnet}. Our proposal is able to directly handle network outputs not limited to 4-wall rooms. Here we demonstrate that is possible to train strategically in a way that the network takes full advantage of the few different data that we have at our disposal.

\textbf{3D Layout Evaluation.}
We evaluate our layout recovering approach on three standard metrics, 3D intersection over union (3DIoU), corner error (CE) and pixel error (PE), and compare ourselves with three works from the state of the art \cite{PanoContext,zou2018layoutnet,fernandez2018layouts}. Results are averaged across all images. 
For all experiments, only SUN360 dataset is used for training.
Table \ref{sun360res} shows the performance of our proposal testing on both datasets. We show that results on SUN360 dataset demonstrate better performance as the FCN has been trained on the same dataset, however results on Standford 2D-3D dataset are also very competitive.

\setlength{\tabcolsep}{4pt}
\begin{table}[t!]
\begin{center}
\caption{Performance benchmarking for SUN360 and Stanford 2D-3D datasets training on SUN360 data. \textit{SS}: Simple Segmentation (3 categories): ceiling, floor and walls \cite{zou2018layoutnet}. \textit{CS}: Complete Segmentation: ceiling, floor, wall$_{1}$,..., wall$_{n}$ \cite{fernandez2018layouts}.}
\label{sun360res}
\scriptsize 
\begin{tabular}{cccccc}

\hline\noalign{\smallskip}
\textbf{Dataset}&\textbf{Method} &\textbf{ 3DIoU(\%)} &\textbf{ CE(\%)} & \textbf{PE$^{SS}$(\%)}& \textbf{PE$^{CS}$(\%)}\\
\noalign{\smallskip}
\hline
\noalign{\smallskip}
SUN360&PanoContext \cite{PanoContext} & 67.22 & 1.60  & 4.55&10.34\\
&F-L C. \textit{et al.} \cite{fernandez2018layouts} & - & - & - & 7.26\\
&LayoutNet \cite{zou2018layoutnet} & 74.48 & 1.06 & 3.34 & -\\
&\textbf{Ours}  & \textbf{76.82} & \textbf{0.79} & \textbf{2.59}& \textbf{3.13}  \\ 
\noalign{\smallskip}
\hline
\noalign{\smallskip}
Stnfd.2D-3D&F-L C. \textit{et al.} \cite{fernandez2018layouts} & - & - & - & 12.1\\
&\textbf{Ours}  & \textbf{70.64} & \textbf{1.15} & \textbf{3.95}& \textbf{4.98} \\
\noalign{\smallskip}
\hline
\end{tabular}
\end{center}
\end{table}
\setlength{\tabcolsep}{1.4pt}

\begin{figure}[t!]
\centering
\subfloat{\includegraphics[width=1\linewidth]{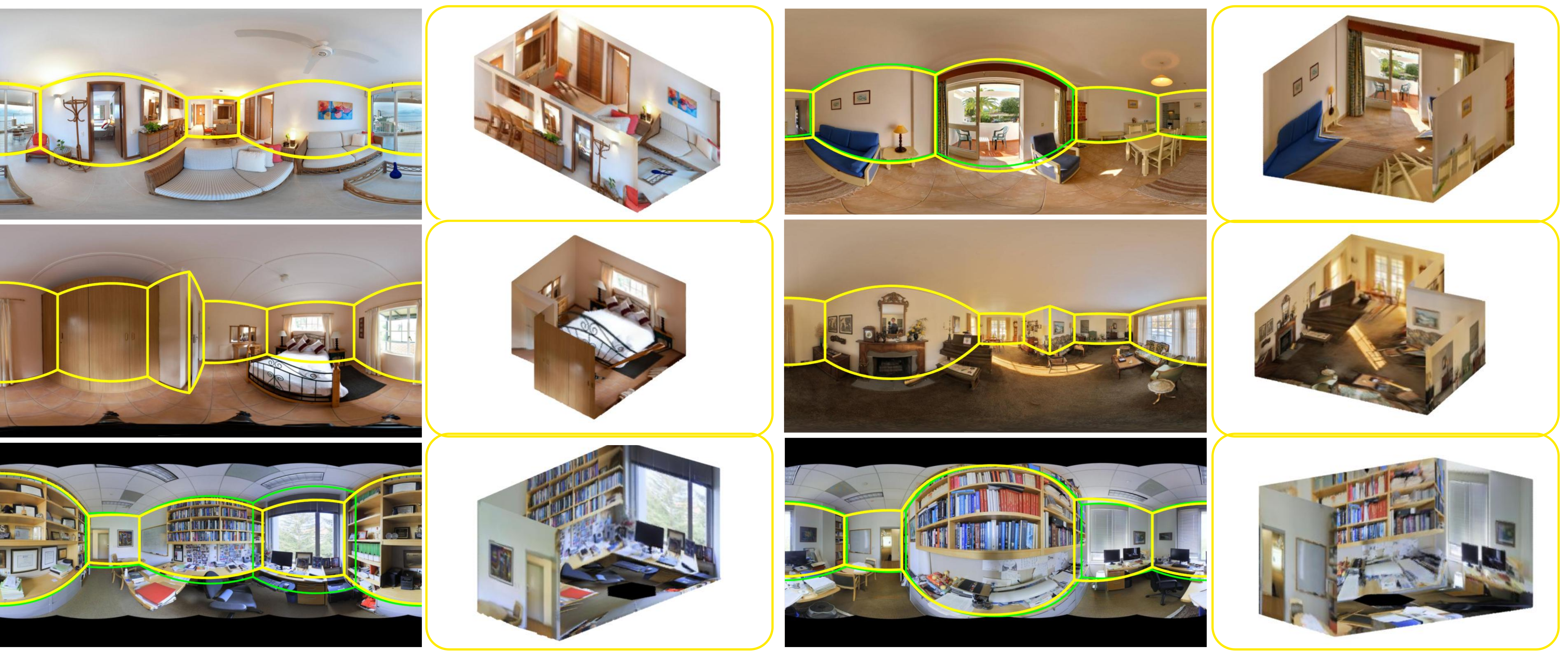}}
\caption{Qualitative results on both datasets. Yellow: PanoRoom, green: GT.}
\end{figure}

\bibliographystyle{splncs}
\bibliography{egbib}
\end{document}